# Arithmetic and Frequency Filtering Methods of Pixel-Based Image Fusion Techniques

Firouz Abdullah Al-Wassai[1], N.V. Kalyankar[2], Ali A. Al-Zuky[3]

[1] Research Student, Computer Science Dept., Yeshwant College, (SRTMU), Nanded, India
fairozwaseai@yahoo.com
[2] Principal, Yeshwant Mahavidyala College, Nanded, India
drkalyankarnv@rediffmail.com
[3] Assistant Professor, Dept.of Physics, College of Science, Mustansiriyah Un. Baghdad – Iraq.
dr.alialzuky@yahoo.com

**Abstract**
In remote sensing, image fusion technique is a useful tool used to fuse high spatial resolution panchromatic images (PAN) with lower spatial resolution multispectral images (MS) to create a high spatial resolution multispectral of image fusion (F) while preserving the spectral information in the multispectral image (MS).There are many PAN sharpening techniques or Pixel-Based image fusion techniques that have been developed to try to enhance the spatial resolution and the spectral property preservation of the MS. This paper attempts to undertake the study of image fusion, by using two types of pixel –based image fusion techniques i.e. Arithmetic Combination and Frequency Filtering Methods of Pixel-Based Image Fusion Techniques. The first type includes Brovey Transform (BT), Color Normalized Transformation (CN) and Multiplicative Method (MLT). The second type include High-Pass Filter Additive Method (HPFA), High – Frequency- Addition Method (HFA) High Frequency Modulation Method (HFM) and The Wavelet transform-based fusion method (WT). This paper also devotes to concentrate on the analytical techniques for evaluating the quality of image fusion (F) by using various methods including Standard Deviation (SD), Entropy(En), Correlation Coefficient (CC), Signal-to Noise Ratio (SNR), Normalization Root Mean Square Error (NRMSE) and Deviation Index (DI) to estimate the quality and degree of information improvement of a fused image quantitatively.

## 1. INTRODUCTION

Although Satellites remote sensing image fusion has been a hot research topic of remote sensing image processing [1]. This is obvious from the amount of conferences and workshops focusing on data fusion, as well as the special issues of scientific journals dedicated to the topic. Previously, data fusion, and in particular image fusion belonged to the world of research and development. In the meantime, it has become a valuable technique for data enhancement in many applications. More and more data providers envisage the marketing of fused products. Software vendors started to offer pre-defined fusion methods within their generic image processing packages [2].

Remote sensing offers a wide variety of image data with different characteristics in terms of temporal, spatial, radiometric and Spectral resolutions. Although the information content of these images might be partially overlapping [3], imaging systems somehow offer a tradeoff between high spatial and high spectral resolution, whereas no single system offers both. Hence, in the remote sensing community, an image with 'greater quality' often means higher spatial or higher spectral resolution, which can only be obtained by more advanced sensors [4]. However, many applications of satellite images require both spectral and spatial resolution to be high. In order to automate the processing of these satellite images new concepts for sensor fusion are needed. It is, therefore, necessary and very useful to be able to merge images with higher spectral information and higher spatial information [5].

The term "fusion" gets several words to appear, such as merging, combination, synergy, integration … and several others that express more or less the same concept have since appeared in literature [6]. Different definitions of data fusion can be found in literature, each author interprets this term differently depending his research interests, such as [7-8] . A general definition of data fusion can be adopted as fallowing "Data fusion is a formal framework which expresses means and tools for the alliance of data originating from different sources. It aims at obtaining information of greater quality; the exact definition of 'greater quality' will depend upon the application" [11-13]. Image fusion forms a subgroup within this definition and aims at the generation of a single image from multiple image data for the extraction of information of higher quality. Having that in mind, the achievement of high spatial resolution, while maintaining the provided spectral resolution, falls exactly into this framework [14].

## 2. Pixel-Based Image Fusion Techniques

Image fusion is a sub area of the more general topic of data fusion [15]. Generally, Image fusion techniques can be classified into three categories depending on the stage at which fusion takes place; it is often divided into three levels, namely: pixel level, feature level and decision level of representation [16, 17] . This paper will focus on pixel level image fusion. The pixel image fusion techniques can be grouped into several techniques depending on the tools or the processing methods for image fusion procedure. It is grouped into four classes: 1) Arithmetic Combination techniques (AC) 2) Component Substitution fusion techniques (CS) 3) Frequency Filtering Methods (FFM) 4) Statistical Methods (SM). This paper focuses on using tow types of pixel –based image fusion techniques Arithmetic Combination and Frequency Filtering Methods of Pixel-Based Image Fusion Techniques. The first type is included BT; CN; MLT and the last type includes HPFA; HFA HFM and WT. In this work to achieve the fusion algorithm and estimate the quality and degree of information improvement of a fused image quantitatively used programming in VB.

To explain the algorithms through this report, Pixels should have the same spatial resolution from two different sources that are manipulated to obtain the resultant image. So, before fusing two sources at a pixel level, it is necessary to perform a geometric registration and a radiometric adjustment of the images to one another. When images are obtained from sensors of different satellites as in the case of fusion of SPOT or IRS with Landsat, the registration accuracy is very important. But registration is not much of a problem with simultaneously acquired images as in the case of Ikonos/Quickbird PAN and MS images. The PAN images have a different spatial resolution from that of MS images. Therefore, resampling of MS images to the spatial resolution of PAN is an essential step in some fusion methods to bring the MS images to the same size of PAN, , thus the resampled MS images will be noted by $M_k$ that represents the set of DN of band k in the resampled MS image . Also the following notations will be used: P as DN for PAN image, $F_k$ the DN in final fusion result for band k. $\bar{M}_k$ $\bar{P}$ , $\sigma_P, \sigma_{M_k}$ Denotes the local means and standard deviation calculated inside the window of size (3, 3) for $M_k$ and P respectively.

## 3. The AC Methods

This category includes simple arithmetic techniques. Different arithmetic combinations have been employed for fusing MS and PAN images. They directly perform some type of arithmetic operation on the MS and PAN bands such as addition, multiplication, normalized division, ratios and subtraction which have been combined in different ways to achieve a better fusion effect. These models assume that there is high correlation between the PAN and each of the MS bands [24]. Some of the popular AC methods for pan sharpening are the BT, CN and MLM. The algorithms are described in the following sections.

### 3.1 Brovey Transform (BT)

The BT, named after its author, uses ratios to sharpen the MS image in this method [18]. It was created to produce RGB images, and therefore only three bands at a time can be merged [19]. Many researchers used the BT to fuse a RGB image with a high resolution image [20-25].The basic procedure of the BT first multiplies each MS band by the high resolution PAN band, and then divides each product by the sum of the MS bands. The following equation, given by [18], gives the mathematical formula for the BT:

$$F_{k(i,j)} = \frac{M_{k(i,j)} \times P_{(i,j)}}{\sum_k M_{k(i,j)}} \quad (1)$$

The BT may cause color distortion if the spectral range of the intensity image is different from the spectral range covered by the MS bands.

### 3.2 Color Normalized Transformation (CN)

CN is an extension of the BT [17]. CN transform also referred to as an energy subdivision transform [26]. The CN transform separates the spectral space into hue and brightness components. The transform multiplies each of the MS bands by the p imagery, and these resulting values are each normalized by being divided by the sum of the MS bands. The CN transform is defined by the following equation [26, 27]:

$$F_{k(i,j)} = \frac{(M_{k(i,j)} + 1.0)(P_{(i,j)} + 1.0) \times 3.0}{\sum_k M_{k(i,j)} + 3.0} - 1.0 \quad (2)$$

(Note: The small additive constants in the equation are included to avoid division by zero.)

### 3.3 Multiplicative Method (MLT)

The Multiplicative model or the product fusion method combines two data sets by multiplying each pixel in each band k of MS data by the corresponding pixel of the PAN data. To compensate for the increased brightness, the square root of the mixed data set is taken. The square root of the Multiplicative data set, reduce the data to combination reflecting the mixed spectral properties of both sets. The fusion algorithm formula is as follows [1; 19 ; 20]:

$$F_{k(i,j)} = \sqrt{M_{k(i,j)} \times P_{(i,j)}} \quad (3)$$

## 4. Frequency Filtering Methods (FFM)

Many authors have found fusion methods in the spatial domain (high frequency inserting procedures) superior over the other approaches, which are known to deliver fusion results that are spectrally distorted to some degree [28] Examples of those authors are [29-31].

Fusion techniques in this group use high pass filters, Fourier transform or wavelet transform to model the frequency components between the PAN and MS images by injecting spatial details in the PAN and introducing them into the MS image. Therefore, the original spectral information of the MS channels is not or only minimally affected [32]. Such algorithms make use of classical filter techniques in the spatial domain. Some of the popular FFM for pan sharpening are the HPF, HFA, HFM and the WT based methods.

### 4.1 High-Pass Filter Additive Method (HPFA)

The High-Pass Filter Additive (HPFA) technique [28] was first introduced by Schowengerdt (1980) as a method to reduce data quantity and increase spatial resolution for Landsat MSS data [33]. HPF basically consists of an addition of spatial details, taken from a high-resolution Pan observation, into the low resolution MS image [34]. The high frequencies information is computed by filtering the PAN with a high-pass filter through a simple local pixel averaging, i.e. box filters. It is performed by emphasize the detailed high frequency components of an image and deemphasize the more general low frequency information [35]. The HPF method uses standard square box HP filters. For example, a 3*3 pixel kernel given by [36], which is used in this study:

$$P_{HPF} = \frac{1}{9}\begin{bmatrix} -1 & -1 & -1 \\ -1 & 8 & -1 \\ -1 & -1 & -1 \end{bmatrix} \quad (4)$$

In its simplest form, The HP filter matrix is occupied by "-1" at all but at the center location. The center value is derived by $c = n * n - 1$, where $c$ is the center value and $n * n$ is the size of the filter box [28]. The HP are filters that comput a local average around each pixel in the PAN image.

The extracted high frequency components of $P_{HPF}$ superimposed on the MS image [1] by simple addition and the result divided by two to offset the increase in brightness values [33]. This technique can improve spatial resolution for either colour composites or an individual band [16]. This is given by [33]:

$$F_k = \frac{(M_k + P_{HPF})}{2} \quad (5)$$

The high frequency is introduced equally without taking into account the relationship between the MS and PAN images. So the HPF alone will accentuate edges in the result but loses a large portion of the information by filtering out the low spatial frequency components [37].

### 4.2 High –Frequency- Addition Method (HFA)

High-frequency-addition method [32] is a technique of filter techniques in spatial domain similar the previous technique, but the difference between them is the way how to extract the high frequencies. In this method, to extract the PAN channel high frequencies; a degraded or low-pass-filtered version of the panchromatic channel has to be created by applying the following set of filter weights (in a 3 x 3 convolution filter example) [38]:

$$P_{LPF} = \frac{1}{9}\begin{bmatrix} 1 & 1 & 1 \\ 1 & 1 & 1 \\ 1 & 1 & 1 \end{bmatrix} \quad (6)$$

A low pass or smoothing filter, which corresponds to computing a local average around each pixel in the image, is achieved. Since the goal of contrast enhancement is to increase the visibility of small detail in an image, subsequently, the high frequency addition method (HFA) extracts the high frequencies using a subtraction procedure .This approach is known as Unsharp masking USM [39]:

$$P_{USM} = P - P_{LPF} \quad (7)$$

Some authors, for example [40]; defined USM as HPF; while [36, 41] multiply the original image by an implication factor, denoted by a, and hence define it as a High Boost Filter (HBF) or high-frequency-emphasis filter: in the original, that is:

$$HBF = a * P - P_{LPF} \quad (8)$$

The general process by using equation (8) called unsharp masking [36] and adds them to the MS channels via addition as shown by equation [32]:

$$F_k = M_k + P_{USM} \quad (9)$$

When this technique is applied, it really leads to the enhancement of all high spatial frequency detail in an image including edges, line and points of high gradient [42]

### 4.3 High Frequency Modulation Method (HFM)

The problem of the addition operation is that the introduced texture will be of different size relative to each multispectral channel, so a channel wise scaling factor for the high frequencies is needed. The alternative high frequency modulation method HFM extracts the high frequencies via division for the P on the PAN channel low frequency $P_{LPF}$ which is obtained by using equation (9) to extract the PAN channel low-frequency $P_{LPF}$ and then adds them to each multispectral channel via multiplication [32]:

$$\boldsymbol{F_k = M_k \times \frac{P}{P_{LPF}}} \quad (10)$$

Because of the multiplication operation, every multispectral channel is modulated by the same high frequencies [32].

### 4.4 Wavelet Transformation (WT) Based Image Fusion

Wavelet-based methods Multi-resolution or multi-scale methods [24] is a mathematical tool developed in the field of signal processing [9] have been adopted for data fusion since the early 1980s (MALAT, 1989). Recently, the wavelet transform approach has been used for fusing data and becomes hot topic in research [43]. The wavelet transform provides a framework to decompose (also called analysis) images into a number of new images, each one of them with a different degree of resolution as well as a perfect reconstruction of the signal (also called synthesis). Wavelet-based approaches show some favorable properties compared to the Fourier transform [44]. While the Fourier transform gives an idea of the frequency content in the image, the wavelet representation is an intermediate representation between the Fourier and the spatial representation, and it can provide good localization in both frequency and space domains [45]. Furthermore, the multi-resolution nature of the wavelet transforms allows for control of fusion quality by controlling the number of resolutions [46] as will as the wavelet transform does not operate on color images directly so we have transformed the color image from RGB domain to anther domain [47].

For more information about image fusion based on wavelet transform have been published in recent years [48 -50].

The block diagram of a generic wavelet-based image fusion scheme is shown in Fig. 3. Wavelet transform based image fusion involves three steps; forward transform coefficient combination and backward transform. In the forward transform, two or more registered input images are wavelet transformed to get their wavelet coefficients [51]. The wavelet coefficients for each level contain the spatial (detail) differences between two successive resolution levels [9].

The basic operation for calculating the DWT is convolving the samples of the input with the low-pass and high-pass filters of the wavelet and down sampling the output [52]. Wavelet transform based image fusion involves various steps:

Step (1): the PAN image $\boldsymbol{P}$ is first reference stretched three times, each time to match one of multispectral $M_k$ histograms to produce three new PAN images.

Step (2): the wavelet basis for the transform is chosen. In this study the upper procedure is for one level wavelet decomposition, and we used to implement the image fusion using wavelet basis of Haar because it is found that the choice of the wavelet basis does affect the fused images [53]. The Haar basis vectors are simple [37]:

$$L = \tfrac{1}{\sqrt{2}}[1 \quad 1] \qquad H = \tfrac{1}{\sqrt{2}}[1 \quad -1] \quad (10)$$

Then performing the wavelet decomposition analysis to extract The structures or "details" present between the images of two different resolution. These structures are isolated into three wavelet coefficients, which correspond to the detailed images according to the three directions. The decomposition at first level we will have one approximation coefficients, ($A^N$ R,G,B) and 3N wavelets Planes for each band by the fallowing equation [54]:

$$R \xrightarrow{WT} A_R^N + \sum_l^N (H_R^l + V_R^l + D_R^l)$$

$$G \xrightarrow{WT} A_G^N + \sum_l^N (H_G^l + V_G^l + D_G^l)$$

$$B \xrightarrow{WT} A_B^N + \sum_l^N (H_B^l + V_B^l + D_B^l) \quad (11)$$

$A^N$: is Approximation coefficient at level N or approximation plane
- $H^l$ : is Horizontal coefficient at level l or horizontal wavelet plane
- $V^l$ : is Vertical Coefficient at level l or vertical wavelet plane
- $D^l$: is Diagonal coefficient at level l or diagonal wavelet plane

Step (3): Similarly by decomposing the panchromatic high-resolution image we will have one approximation coefficients, $(A_P^N)$ and *3N* wavelets Planes for Panchromatic image, where *PAN* means, panchromatic image.

Step (4): the wavelet coefficients sets from two images are combined via substitutive or additive rules. In the case of substitutive method, the wavelet coefficient planes (or details) of the *R, G*, and *B* decompositions are replaced by the similar detail planes of the panchromatic decomposition, which that used in this study.

Step (5): Then, for obtaining the fused images, the inverse wavelet transform is implemented on resultant sets. By reversing the process in step (2) the synthesis is equation [54]:

$$A_R^N + \sum_l^N (H_P^l + V_P^l + D_P^l) \xrightarrow{IWT} R_{new}$$
$$A_G^N + \sum_l^N (H_P^l + V_P^l + D_P^l) \xrightarrow{IWT} G_{new}$$
$$A_B^N + \sum_l^N (H_P^l + V_P^l + D_P^l) \xrightarrow{IWT} B_{new} \quad (12)$$

Wavelet transform fusion is obtained. This reverse process is referred to as reconstruction of the image in which the finer representation is calculated from coarser levels by adding the details according to the synthesis equation [44]. Thus at high resolution, simulated are produced.

## 5. Experiments

In order to validate the theoretical analysis, the performance of the methods discussed above was further evaluated by experimentation. Data sets used for this study were collected by the Indian IRS-1C PAN (0.50 - 0.75 µm) of the 5.8- m resolution panchromatic band. Where the American Landsat (TM) the red (0.63 - 0.69 µm), green (0.52 - 0.60 µm) and blue (0.45 - 0.52 µm) bands of the 30 m resolution multispectral image were used in this experiment. Fig. 3 shows the IRS-1C PAN and multispectral TM images. The scenes covered the same area of the Mausoleums of the Chinese Tang – Dynasty in the PR China [55] was selected as test sit in this study. Since this study is involved in evaluation of the effect of the various spatial, radiometric and spectral resolution for image fusion, an area contains both manmade and natural features is essential to study these effects. Hence, this work is an attempt to study the quality of the images fused from different sensors with various characteristics. The size of the PAN is 600 * 525 pixels at 6 bits per pixel and the size of the original multispectral is 120 * 105 pixels at 8 bits per pixel, but this is upsampled to by Nearest neighbor was used to avoid spectral contamination caused by interpolation.

To evaluate the ability of enhancing spatial details and preserving spectral information, some Indices including Standard Deviation (SD), Entropy(En), Correlation Coefficient (CC), Signal-to Noise Ratio (SNR), Normalization Root Mean Square Error (NRMSE) and Deviation Index (DI) of the image were used (Table 1), and the results are shown in Table 2. In the following sections, $F_k$, $M_k$ are the measurements of each the brightness values of homogenous pixels of the result image and the original multispectral image of band k, $\overline{M}_k$ and $\overline{F}_k$ are the mean brightness values of both images and are of size $n * m$. BV is the brightness value of image data $\overline{M}_k$ and $\overline{F}_k$. To simplify the comparison of the different fusion methods, the values of the En, CC, SNR, NRMSE and DI index of the fused images are provided as chart in Fig. 1

Table 1: Indices Used to Assess Fusion Images.

| Equation |
|---|
| $\sigma_k = \sqrt{\dfrac{\sum_{i=1}^{m}\sum_{j=1}^{n}(BV_k(n,m)-\mu_k)^2}{m \times n}}$ |
| $CC_k = \dfrac{\sum_i^n \sum_j^m (F_k(i,j) - \overline{F}_k)(M_k(i,j) - \overline{M}_k)}{\sqrt{\sum_i^n \sum_j^m (F_k(i,j) - \overline{F}_k)^2} \sqrt{\sum_i^n \sum_j^m (M_k(i,j) - \overline{M}_k)^2}}$ |
| $En = - \sum_{0}^{I-1} P(i) \log_2 P(i)$ |
| $DI_k = \dfrac{1}{nm}\sum_i^n \sum_j^m \dfrac{|F_k(i,j) - M_k(i,j)|}{M_k(i,j)}$ |
| $SNR_k = \sqrt{\dfrac{\sum_i^n \sum_j^m (F_k(i,j))^2}{\sum_i^n \sum_j^m (F_k(i,j) - M_k(i,j))^2}}$ |
| $NRMSE_k = \sqrt{\dfrac{1}{nm * 255^2} \sum_i^n \sum_j^m (F_k(i,j) - M_k(i,j))^2}$ |

## 6. Discussion Of Results

The Fig. 1 shows those parameters for the fused images using various methods. It can be seen that from fig.1a. The SD of the fused images remains constant for HFA and HFM. According to the computation results En, the increased En indicates the change in quantity of information content for radiometric resolution through the merging. From fig.1b, it is obvious that En of the fused images have been changed when compared to the original multispectral but some methods such as (BT and HPFA) decrease the En values to below the original. In Fig.1c.Correlation values also remain practically constant, very near the maximum possible value except BT and CN. The results of SNR, NRMSE and DI appear changing significantly. It can be observed, from the diagram of Fig. 1., that the results of NRMSE & DI, of the fused image, show that the HFM and HFA methods give the best results with respect to the other methods indicating that these methods maintain most of information spectral content of the original multispectral data set which get the same values presented the lowest value of the NRMSE & DI as well as the higher of the SNR. Hence, the spectral qualities of fused images by HFM and HFA methods are much better than the others. In contrast, it can also be noted that the BT, HPFA images produce highly NRMSE & DI values indicate that these methods deteriorate spectral information content for the reference image. In a comparison of spatial effects, it can be seen that the results of the HFM; HFA; WT and CN are better than other methods. Fig.3. shows the original images and the fused image results.

By combining the visual inspection results, it can be seen that the experimental results overall method are The HFM and HFA results which are the best result. The next higher the visual inspection results are obtained with WT, CN and MUL.

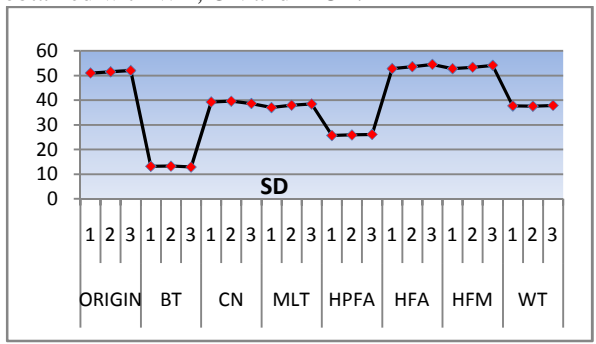

Fig. 1a: Chart Representation of SD of Fused Images

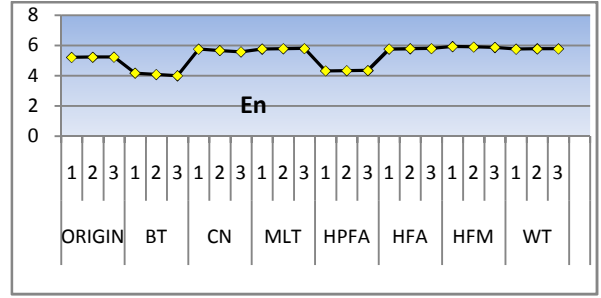

Fig. 1b: Chart Representation of En of Fused Images

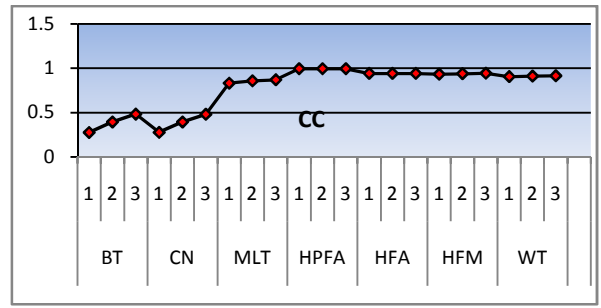

Fig. 1c: Chart Representation of CC of Fused Images

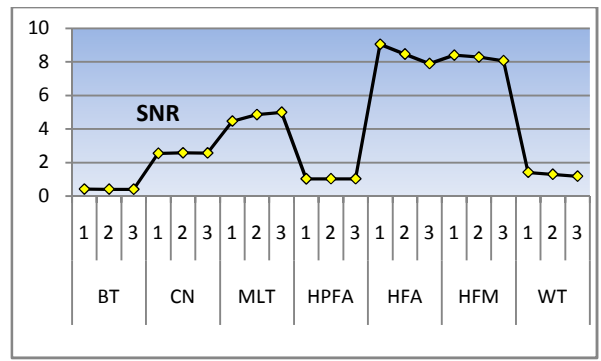

Fig. 1d: Chart Representation of SNR of Fused Images

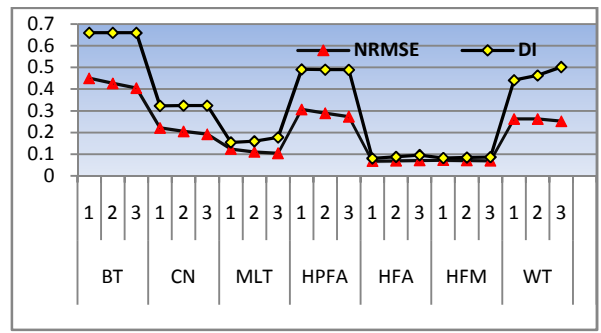

Fig. 1e: Chart Representation of NRMSE & DI of Fused Images

Fig. 1: Chart Representation of SD , En , CC ,NRMSE & DI of Fused Images

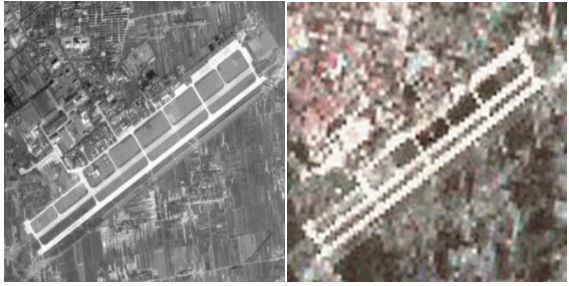

Fig.2a. Original Panchromatic    Fig.2b. Original Multispectral

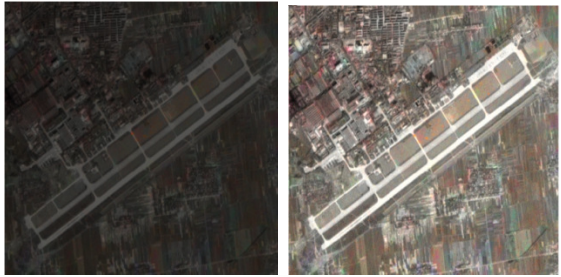

Fig. 2c. BT    Fig.2d. CN

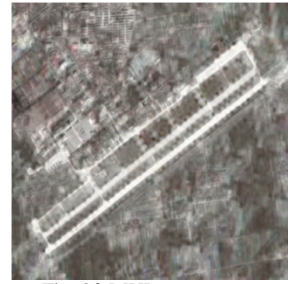

Fig. 2f. MUL

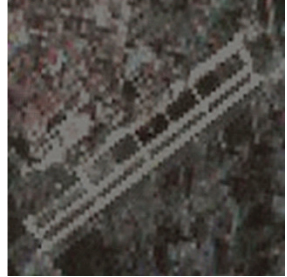

Fig. 2g. HPF

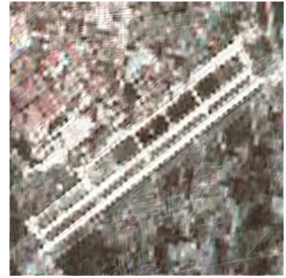

Fig. 2e. HFA

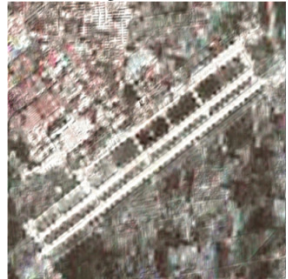

Fig. 2f. HFM

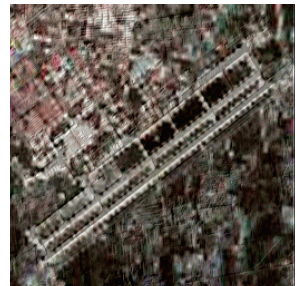

**Fig. 2i. WT**
Fig.2: The Representation of orginal and Fused Images

Table 2: Quantitative Analysis of Original MS and Fused Image Results Through the Different Methods

| Method | Band | SD | En | SNR | NRMSE | DI | CC |
|---|---|---|---|---|---|---|---|
| ORIGIN | 1 | 51.018 | 5.2093 | | | | |
| | 2 | 51.477 | 5.2263 | | | | |
| | 3 | 51.983 | 5.2326 | | | | |
| BT | 1 | 13.185 | 4.1707 | 0.416 | 0.45 | 0.66 | 0.274 |
| | 2 | 13.204 | 4.0821 | 0.413 | 0.427 | 0.66 | 0.393 |
| | 3 | 12.878 | 3.9963 | 0.406 | 0.405 | 0.66 | 0.482 |
| CN | 1 | 39.278 | 5.7552 | 2.547 | 0.221 | 0.323 | 0.276 |
| | 2 | 39.589 | 5.6629 | 2.579 | 0.205 | 0.324 | 0.393 |
| | 3 | 38.633 | 5.5767 | 2.57 | 0.192 | 0.324 | 0.481 |
| MLT | 1 | 37.009 | 5.7651 | 4.468 | 0.124 | 0.154 | 0.832 |
| | 2 | 37.949 | 5.7833 | 4.858 | 0.111 | 0.159 | 0.859 |
| | 3 | 38.444 | 5.7915 | 4.998 | 0.104 | 0.177 | 0.871 |
| HPFA | 1 | 25.667 | 4.3176 | 1.03 | 0.306 | 0.491 | 0.996 |
| | 2 | 25.869 | 4.3331 | 1.032 | 0.289 | 0.49 | 0.996 |
| | 3 | 26.121 | 4.3424 | 1.033 | 0.273 | 0.489 | 0.996 |
| HFA | 1 | 52.793 | 5.7651 | 9.05 | 0.068 | 0.08 | 0.943 |
| | 2 | 53.57 | 5.7833 | 8.466 | 0.07 | 0.087 | 0.943 |
| | 3 | 54.498 | 5.7915 | 7.9 | 0.071 | 0.095 | 0.943 |
| HFM | 1 | 52.76 | 5.9259 | 8.399 | 0.073 | 0.082 | 0.934 |
| | 2 | 53.343 | 5.8979 | 8.286 | 0.071 | 0.084 | 0.94 |
| | 3 | 54.136 | 5.8721 | 8.073 | 0.069 | 0.086 | 0.945 |
| WT | 1 | 37.666 | 5.7576 | 1.417 | 0.262 | 0.441 | 0.907 |
| | 2 | 37.554 | 5.7754 | 1.296 | 0.262 | 0.463 | 0.913 |
| | 3 | 37.875 | 5.7765 | 1.182 | 0.252 | 0.502 | 0.916 |

## 6. Conclusion

Image Fusion aims at the integration of disparate and complementary data to enhance the information apparent in the images as well as to increase the reliability of the interpretation. This leads to more accurate data and increased utility in application fields like segmentation and classification. In this paper, the comparative studies undertaken by using two types of pixel –based image fusion techniques Arithmetic Combination and Frequency Filtering Methods of Pixel-Based Image Fusion Techniques as well as effectiveness based image fusion and the performance of these methods. The fusion procedures of the first type, which includes (BT; CN; MLT ) by using all PAN band, produce more distortion of spectral characteristics because such methods depend on the degree of global correlation between the PAN and multispectral bands to be enhanced. Therefore, these fusion techniques are not adequate to preserve the spectral characteristics of original multispectral. But those methods enhance the spatial quality of the imagery except BT. The fusion procedures of the second type includes HPFA; HFA; HFM and the WT based fusion method by using selected (or Filtering) PAN band frequencies including HPF, HFA, HFM and WT algorithms. The preceding analysis shows that the HFA and HFM methods maintain the spectral integrity and enhance the spatial quality of the imagery. The HPF method does not maintain the spectral integrity and does not enhance the spatial quality of the imagery. The WTF method has been shown in many published papers as an efficient image fusion. In the present work, the WTF method has shown low results.

In general types of the data fusion techniques, the use of the HFM &HFA could, therefore, be strongly recommended if the goal of the merging is to achieve the best representation of the spectral information of multispectral image and the spatial details of a high-resolution panchromatic image.

## References

[1] Wenbo W.,Y.Jing, K. Tingjun ,2008. "Study Of Remote Sensing Image Fusion And Its Application In Image Classification" The International Archives of the Photogrammetry, Remote Sensing and Spatial Information Sciences. Vol. XXXVII. Part B7. Beijing 2008, pp.1141-1146.

[2] Pohl C., H. Touron, 1999. "Operational Applications of Multi-Sensor Image Fusion". International Archives of Photogrammetry and Remote Sensing, Vol. 32, Part 7-4-3 w6, Valladolid, Spain.

[3] Steinnocher K., 1999. "Adaptive Fusion Of Multisource Raster Data Applying Filter Techniques". International Archives of Photogrammetry and Remote Sensing, Vol. 32, Part 7-4-3 W6, Valladolid, Spain, 3-4 June, pp.108-115.

[4] Dou W., Chen Y., Li W., Daniel Z. Sui, 2007. "A General Framework for Component Substitution Image Fusion: An Implementation Using the Fast Image Fusion Method". Computers & Geosciences 33 (2007), pp. 219–228.

[5] Zhang Y., 2004."Understanding Image Fusion". Photogrammetric Engineering & Remote Sensing, pp. 657-661.

[6] Wald L., 1999a, "Some Terms Of Reference In Data Fusion". IEEE Transactions on Geosciences and Remote Sensing, 37, 3, pp.1190- 1193.

[7] Hall D. L. and Llinas J., 1997. "An introduction to multisensor data fusion," (invited paper) in Proceedings of the IEEE, Vol. 85, No 1, pp. 6-23.

[8] Pohl C. and Van Genderen J. L., 1998. "Multisensor Image Fusion In Remote Sensing: Concepts, Methods And Applications".(Review Article), International Journal Of Remote Sensing, Vol. 19, No.5, pp. 823-854.

[9] Zhang Y., 2002. "PERFORMANCE ANALYSIS OF IMAGE FUSION TECHNIQUES BY IMAGE". International Archives of Photogrammetry and Remote Sensing (IAPRS), Vol. 34, Part 4. Working Group IV/7.

[11] Ranchin, T., L. Wald, M. Mangolini, 1996a, "The ARSIS method: A General Solution For Improving Spatial Resolution Of Images By The Means Of Sensor Fusion". Fusion of Earth Data, Proceedings EARSeL Conference, Cannes, France, 6- 8 February 1996(Paris: European Space Agency).

[12] Ranchin T., L.Wald , M. Mangolini, C. Penicand, 1996b. "On the assessment of merging processes for the improvement of the spatial resolution of multispectral SPOT XS images". In Proceedings of the conference, Cannes, France, February 6-8, 1996, published by SEE/URISCA, Nice, France, pp. 59-67

[13] Wald L., 1999b, "Definitions And Terms Of Reference In Data Fusion". International Archives of Photogrammetry and Remote Sensing, Vol. 32, Part 7-4-3 W6, Valladolid, Spain, 3-4 June.

[14] Pohl C., 1999." Tools And Methods For Fusion Of Images Of Different Spatial Resolution". International Archives of Photogrammetry and Remote Sensing, Vol. 32, Part 7-4-3 W6, Valladolid, Spain, 3-4 June.

[15] Hsu S. H., Gau P. W., I-Lin Wu I., and Jeng J. H., 2009,"Region-Based Image Fusion with Artificial Neural Network". World Academy of Science, Engineering and Technology, 53, pp 156 -159.

[16] Zhang J., 2010. "Multi-source remote sensing data fusion: status and trends", International Journal of Image and Data Fusion, Vol. 1, No. 1, pp. 5–24.

[17] Ehlers M., S. Klonusa, P. Johan A ̊ strand and P. Rosso ,2010. "Multi-sensor image fusion for pansharpening in remote sensing". International Journal of Image and Data Fusion, Vol. 1, No. 1, March 2010, pp. 25–45

[18] Vijayaraj V., O'Hara C. G. And Younan N. H., 2004."Quality Analysis Of Pansharpened Images". 0-7803-8742-2/04/(C) 2004 IEEE,pp.85-88

[19] ŠVab A. and Oštir K., 2006. "High-Resolution Image Fusion: Methods To Preserve Spectral And Spatial Resolution". Photogrammetric Engineering & Remote Sensing, Vol. 72, No. 5, May 2006, pp. 565–572.

[20] Parcharidis I. and L. M. K. Tani, 2000. "Landsat TM and ERS Data Fusion: A Statistical Approach Evaluation for Four Different Methods". 0-7803-6359-0/00/ 2000 IEEE, pp.2120-2122.

[21] Ranchin T., Wald L., 2000. "Fusion of high spatial and spectral resolution images: the ARSIS concept and its implementation". Photogrammetric Engineering and Remote Sensing, Vol.66, No.1, pp.49-61.

[22] Prasad N., S. Saran, S. P. S. Kushwaha and P. S. Roy, 2001. "Evaluation Of Various Image Fusion Techniques And Imaging Scales For Forest Features Interpretation". Current Science, Vol. 81, No. 9, pp.1218

[23] Alparone L., Baronti S., Garzelli A., Nencini F., 2004. " Landsat ETM+ and SAR Image Fusion Based on Generalized Intensity Modulation". IEEE Transactions on Geoscience and Remote Sensing, Vol. 42, No. 12, pp. 2832-2839

[24] Dong J.,Zhuang D., Huang Y.,Jingying Fu,2009. "Advances In Multi-Sensor Data Fusion: Algorithms And Applications ". Review , ISSN 1424-8220 Sensors 2009, 9, pp.7771-7784.

[25] Amarsaikhan D., H.H. Blotevogel, J.L. van Genderen, M. Ganzorig, R. Gantuya and B. Nergui, 2010. "Fusing high-resolution SAR and optical imagery for improved urban land cover study and classification". International Journal of Image and Data Fusion, Vol. 1, No. 1, March 2010, pp. 83–97.

[26] Vrabel J., 1996. "Multispectral imagery band sharpening study". Photogrammetric Engineering and Remote Sensing, Vol. 62, No. 9, pp. 1075-1083.

[27] Vrabel J., 2000. "Multispectral imagery Advanced band sharpening study". Photogrammetric Engineering and Remote Sensing, Vol. 66, No. 1, pp. 73-79.

[28] Gangkofner U. G., P. S. Pradhan, and D. W. Holcomb, 2008. "Optimizing the High-Pass Filter Addition Technique for Image Fusion". Photogrammetric Engineering & Remote Sensing, Vol. 74, No. 9, pp. 1107–1118.

[29] Wald L., T. Ranchin and M. Mangolini, 1997. 'Fusion of satellite images of different spatial resolutions: Assessing the quality of resulting images', Photogrammetric Engineering and Remote Sensing, Vol. 63, No. 6, pp. 691–699.

[30] Li J., 2001. "Spatial Quality Evaluation Of Fusion Of Different Resolution Images". International Archives of Photogrammetry and Remote Sensing. Vol. XXXIII, Part B2, Amsterdam 2000, pp.339-346.

[31] Aiazzi, B., L. Alparone, S. Baronti, I. Pippi, and M. Selva, 2003. "Generalised Laplacian pyramid-based fusion of MS + P image data with spectral distortion minimization".URL:http://www.isprs.org/ commission3/proceedings02/papers/paper083.pdf (Last date accessed: 8 Feb 2010).

[32] Hill J., C. Diemer, O. Stöver, Th. Udelhoven, 1999. "A Local Correlation Approach for the Fusion of Remote Sensing Data with Different Spatial Resolutions in Forestry Applications". International Archives Of Photogrammetry And Remote Sensing, Vol. 32, Part 7-4-3 W6, Valladolid, Spain, 3-4 June.

[33] Carter, D.B., 1998. "Analysis of Multiresolution Data Fusion Techniques". Master Thesis Virginia Polytechnic Institute and State University, URL: http://scholar.lib.vt.edu/theses/available /etd-32198–21323/unrestricted/Etd.pdf (last date accessed: 10 May 2008).

[34] Aiazzi B., S. Baronti , M. Selva,2008. "Image fusion through multiresolution oversampled decompositions". in Image Fusion: Algorithms and Applications ".Edited by: Stathaki T. "Image Fusion: Algorithms and Applications". 2008 Elsevier Ltd.

[35] Lillesand T., and Kiefer R.1994. "Remote Sensing And Image Interpretation". 3rd Edition, John Wiley And Sons Inc.,

[36] Gonzales R. C, and R. Woods, 1992. "Digital Image Processing". A ddison-Wesley Publishing Company.

[37] Umbaugh S. E., 1998. "Computer Vision and Image Processing: Apractical Approach Using CVIP tools". Prentic Hall.

[38] Green W. B., 1989. Digital Image processing A system Approach".2nd Edition. Van Nostrand Reinhold, New York.

[39] Sangwine S. J., and R.E.N. Horne, 1989. "The Colour Image Processing Handbook". Chapman & Hall.

[40] Gross K. and C. Moulds, 1996. Digital Image Processing. (http://www.net/Digital Image Processing.htm). (last date accessed: 10 Jun 2008).

[41] Jensen J.R., 1986. "Introductory Digital Image Processing A Remote Sensing Perspective". Englewood Cliffs, New Jersey: Prentice-Hall.

[42] Richards J. A., and Jia X., 1999. "Remote Sensing Digital Image Analysis". 3rd Edition. Springer - verlag Berlin Heidelberg New York.

[43] Cao D., Q. Yin, and P. Guo,2006. "Mallat Fusion for Multi-Source Remote Sensing Classification". Proceedings of the Sixth International Conference on Intelligent Systems Design and Applications (ISDA'06)

[44] Hahn M. and F. Samadzadegan, 1999. " Integration of DTMS Using Wavelets". International Archives Of Photogrammetry And Remote Sensing, Vol. 32, Part 7-4-3 W6, Valladolid, Spain, 3-4 June. 1999.

[45] King R. L. and Wang J., 2001. "A Wavelet Based Algorithm for Pan Sharpening Landsat 7 Imagery". 0-7803-7031-7/01/ 02001 IEEE, pp. 849- 851

[46] Kumar Y. K.,. "Comparison Of Fusion Techniques Applied To Preclinical Images: Fast Discrete Curvelet Transform Using Wrapping Technique & Wavelet Transform". Journal Of Theoretical And Applied Information Technology.© 2005 - 2009 Jatit, pp. 668-673

[47] Malik N. H., S. Asif M. Gilani, Anwaar-ul-Haq, 2008. "Wavelet Based Exposure Fusion". Proceedings of the World Congress on Engineering 2008 Vol I WCE 2008, July 2 - 4, 2008, London, U.K

[48] Li S., Kwok J. T., Wang Y.., 2002. "Using The Discrete Wavelet Frame Transform To Merge Landsat TM And SPOT Panchromatic Images". Information Fusion 3 (2002), pp.17–23.

[49] Garzelli, A. and Nencini, F., 2006. "Fusion of panchromatic and multispectral images by genetic Algorithms". IEEE Transactions on Geoscience and Remote Sensing, 40, 3810–3813.
[50] Aiazzi, B., Baronti, S., and Selva, M., 2007. "Improving component substitution pan-sharpening through multivariate regression of MS+Pan data". IEEE Transactions on Geoscience and Remote Sensing, Vol.45, No.10, pp. 3230–3239.
[51] Das A. and Revathy K., 2007. "A Comparative Analysis of Image Fusion Techniques for Remote Sensed Images". Proceedings of the World Congress on Engineering 2007, Vol. I, WCE 2007, July 2 – 4,London, U.K.
[52] Pradhan P.S., King R.L., 2006. "Estimation of the Number of Decomposition Levels for a Wavelet-Based Multi-resolution Multi-sensor Image Fusion". IEEE Transaction of Geosciences and Remote Sensing, Vol. 44, No. 12, pp. 3674-3686.
[53] Hu Deyong H. L., 1998. "A fusion Approach of Multi-Sensor Remote Sensing Data Based on Wavelet Transform". URL: http://www.gisdevelopment.net/AARS/ACRS1998/Digital Image Processing (last date accessed: 15 Feb 2009).
[54] Li S.,Li Z.,Gong J.,2010."Multivariate statistical analysis of measures for assessing the quality of image fusion". International Journal of Image and Data Fusion Vol. 1, No. 1, March 2010, pp. 47–66.
[55] Böhler W. and G. Heinz, 1998. "Integration of high Resolution Satellite Images into Archaeological Docmentation". Proceeding International Archives of Photogrammetry and Remote Sensing, Commission V, Working Group V/5, CIPA International Symposium, Published by the Swedish Society for Photogrammetry and Remote Sensing, Goteborg. (URL: http://www.i3mainz.fh-mainz.de/publicat/cipa-98/sat-im.html (Last date accessed: 28 Oct. 2000).

**Authors**

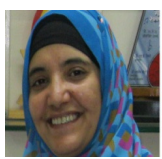

*Firouz Abdullah Al-Wassai*. Received the B.Sc. degree in, Physics from University of Sana'a, Yemen, Sana'a, in 1993. The M.Sc.degree in, Physics from Bagdad University , Iraqe, in 2003, Research student.Ph.D in the department of computer science (S.R.T.M.U), India, Nanded.

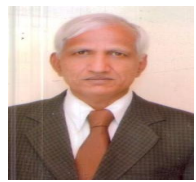

*Dr. N.V. Kalyankar*, Principal,Yeshwant Mahvidyalaya, Nanded(India) completed M.Sc.(Physics) from Dr. B.A.M.U, Aurangabad. In 1980 he joined as a leturer in department of physics at Yeshwant Mahavidyalaya, Nanded. In 1984 he completed his DHE. He completed his Ph.D. from Dr.B.A.M.U. Aurangabad in 1995. From 2003 he is working as a Principal to till date in Yeshwant Mahavidyalaya, Nanded. He is also research guide for Physics and Computer Science in S.R.T.M.U, Nanded. 03 research students are successfully awarded Ph.D in Computer Science under his guidance. 12 research students are successfully awarded M.Phil in Computer Science under his guidance He is also worked on various boides in S.R.T.M.U, Nanded. He is also worked on various bodies is S.R.T.M.U, Nanded. He also published 30 research papers in various international/national journals. He is peer team member of NAAC (National Assessment and Accreditation Council, India ). He published a book entilteld "DBMS concepts and programming in Foxpro". He also get various educational wards in which "Best Principal" award from S.R.T.M.U, Nanded in 2009  and "Best Teacher" award from Govt. of Maharashtra, India in 2010. He is life member of Indian "Fellowship of Linnean Society of London(F.L.S.)" on 11 National Congress, Kolkata (India). He is also honored with November 2009.

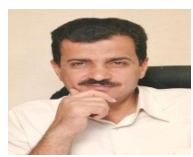

*Dr. Ali A. Al-Zuky*. B.Sc Physics Mustansiriyah University, Baghdad , Iraq, 1990. M Sc. In1993 and Ph. D. in1998 from University of Baghdad, Iraq. He was supervision for 40 postgraduate students (MSc. & Ph.D.) in different fields (physics, computers and Computer Engineering and Medical Physics). He has More than 60 scientific papers published in scientific journals in several scientific conferences.